\documentclass{article}

\usepackage{arxiv}

\usepackage[utf8]{inputenc} 
\usepackage[T1]{fontenc}    
\usepackage{hyperref}       
\usepackage{url}            
\usepackage{booktabs}       
\usepackage{amsfonts}       
\usepackage{nicefrac}       
\usepackage{microtype}      
\usepackage{lipsum}
\usepackage{graphicx}
\usepackage{multirow}
\usepackage{algorithm}        
\usepackage{algpseudocode}    
\usepackage{amsmath}          
\usepackage{bm}  
\usepackage[numbers]{natbib}  
\usepackage{subfig}   
\graphicspath{ {./images/} }

\title{Entropy-Constrained Strategy Optimization in Urban Floods: A Multi-Agent Framework with LLM and Knowledge Graph Integration}

\author{
 Peilin Ji\\
  College of Intelligence and Computing\\
  Tianjin University\\
  Tianjin 300072, China\\
  \texttt{ji\_peilin@tju.edu.cn}
    \And
 Xiao Xue\thanks{Corresponding author. Email: \texttt{jzxuexiao@tju.edu.cn}}\\
  College of Intelligence and Computing\\
  Tianjin University\\
  Tianjin 300072, China\\
  \texttt{jzxuexiao@tju.edu.cn}
  \And
 Simeng Wang\\
  College of Intelligence and Computing\\
  Tianjin University\\
  Tianjin 300072, China\\
  \texttt{2024439163@tju.edu.cn}
  \And
 Wenhao Yan\\
  College of Intelligence and Computing\\
  Tianjin University\\
  Tianjin 300072, China\\
  \texttt{2024439164@tju.edu.cn}
}

\begin{document}
\maketitle
\begin{abstract}
In recent years, the increasing frequency of extreme urban rainfall events has posed significant challenges to emergency scheduling systems. Urban flooding often leads to severe traffic congestion and service disruptions, threatening public safety and mobility. Yet, effective decision-making remains hindered by three key challenges: (1) managing trade-offs among competing goals—such as traffic flow, task completion, and risk mitigation—requires dynamic, context-aware strategies; (2) rapidly evolving environmental conditions render static rules inadequate; and (3) LLM-generated strategies frequently suffer from semantic instability and execution inconsistency. Existing methods fail to align perception, global optimization, and multi-agent coordination within a unified framework. To tackle these challenges, we introduce H–J, a hierarchical multi-agent framework that integrates knowledge-guided prompting, entropy-constrained generation, and feedback-driven optimization. It establishes a closed-loop pipeline spanning from multi-source perception to strategic execution and continuous refinement. We assess H–J on real-world urban topology and rainfall data under three representative conditions: extreme rainfall, intermittent bursts, and daily light rain. Experiments show that H–J significantly outperforms both rule-based and reinforcement learning baselines in traffic smoothness, task success rate, and system robustness. These findings underscore the promise of uncertainty-aware, knowledge-constrained LLM-based frameworks for enhancing resilience in urban flood response.
\end{abstract}


\section{Introduction}
Urban flood events have surged due to intensifying climate patterns, causing significant disruptions to transportation and public service systems~\cite{liang2025_dynamicdispatch, fan2020_adaptiveRLmobility}. Under extreme rainfall, waterlogging and human congestion call for intelligent scheduling systems capable of adaptive coordination to ensure safety and operational continuity.

Yet, flood emergency scheduling remains challenging due to:
(1) the need to dynamically balance competing goals—traffic flow, arrival rate, and risk mitigation~\cite{li2024_reinforcerouting};
(2) rapidly changing environments, where static rules fail and reinforcement learning (RL) methods often struggle with control and generalizability~\cite{emmanuel2025_rl_evacuations};
and (3) semantic‑execution mismatch in LLM‑generated strategies, which lack structural grounding and adaptive correction~\cite{nazareno2025_spatialRAG, xu2025_multimodalLLM}.
Existing solutions typically focus on isolated tasks (e.g., routing or congestion prediction) but fall short of offering a unified, feedback‑aware decision pipeline.

\begin{figure}[h] 
    \centering
    \includegraphics[width=0.65\linewidth, keepaspectratio]{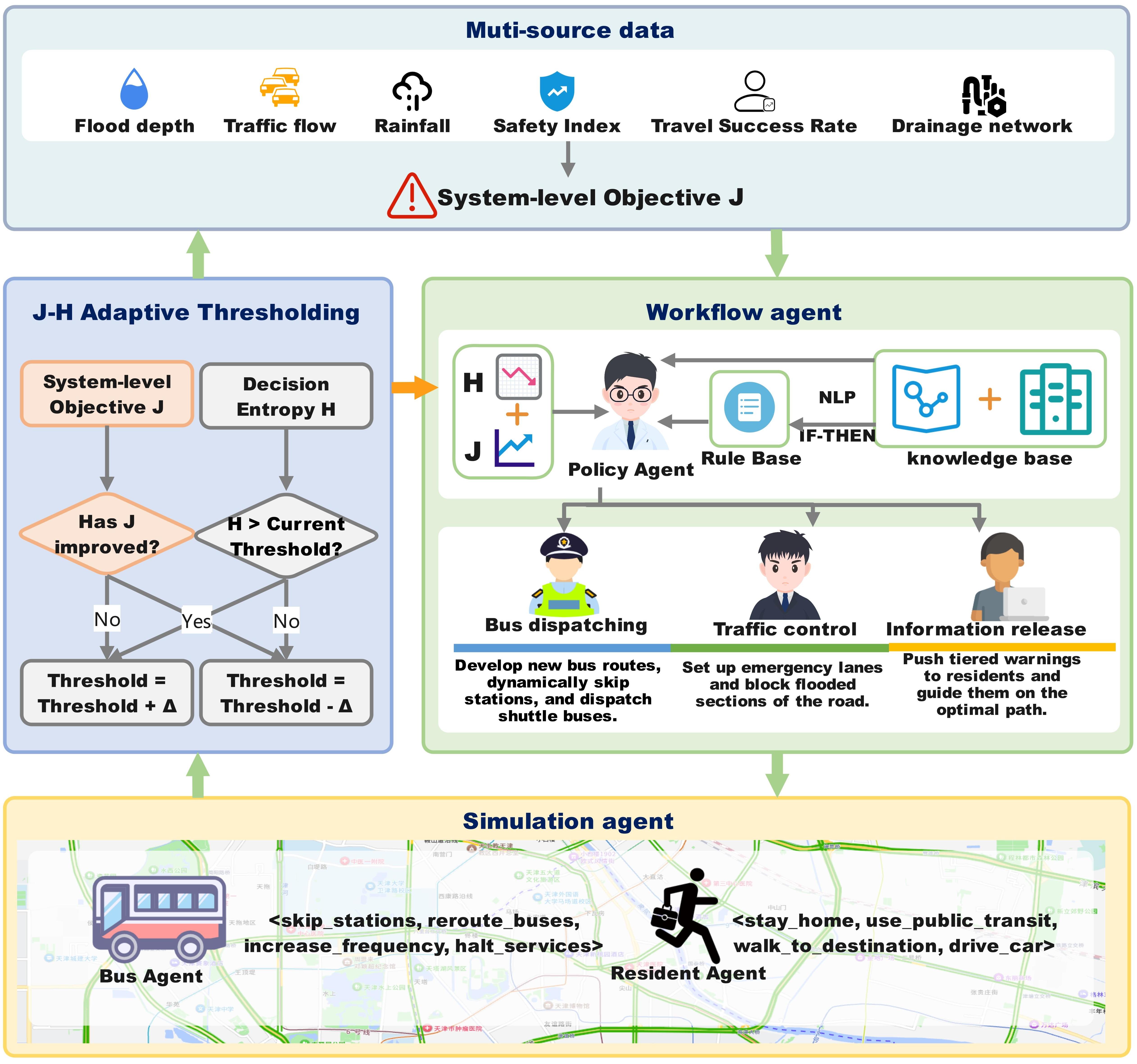}
    \caption{System-level overview of the proposed H--J framework. The architecture integrates multi-source perception, knowledge-guided policy generation, entropy-based threshold adjustment, and simulation-based feedback, forming a closed-loop scheduling cycle for urban flood scenarios.}
  \label{fig:hj-overview}
\end{figure}
To address these issues, we propose the H–J framework (Hierarchical Joint Optimization): a large‑model‑driven, multi‑agent strategy system tailored for urban floods.  H–J unifies four modules—LLMs, dual-channel knowledge retrieval, entropy-constrained generation, and feedback optimization—forming a closed-loop pipeline from perception to adaptive policy refinement. The core innovations include:

\textbf{Dual-channel knowledge retrieval:} A hybrid retrieval mechanism fuses structured graphs and unstructured logs (e.g., dispatch history, road semantics, risk profiles) into context-aware prompts, guiding the LLM toward relevant and constrained strategy generation.

\textbf{Entropy-constrained hierarchical generation:} To enhance stability, we explicitly regulate information entropy across policy layers. High-level planning undergoes entropy compression and categorical filtering to suppress semantic drift, while low-level execution preserves diversity—ensuring robust yet flexible multi-agent strategies.

\textbf{Objective-driven feedback optimization:} A weighted global objective \( J \), integrating flood severity, traffic congestion, task cancellations, and arrival rate, continuously evaluates policy effectiveness and triggers prompt adaptation, forming a closed optimization loop.

We evaluate H–J using real-world rainfall and urban topology data from a representative metropolitan area. Across extreme, intermittent, and light rainfall scenarios, H–J consistently outperforms rule-based and RL baselines in traffic mitigation, task fulfillment, and scenario generalization.
\section{Related Work}

Effective decision-making in urban flood scheduling must address trade-offs among objectives, dynamic uncertainty, and execution feasibility. Prior studies fall into three key directions: reinforcement learning (RL), LLM-based planning, and multi-agent structural modeling. However, none provide a closed-loop, feedback-aware strategy system with semantic and operational robustness.

\subsection{Reinforcement Learning–Based Dispatching Mechanisms}

RL methods such as Proximal Policy Optimization (PPO) have proven effective in traffic signal control and routing~\cite{tan2019DRLsignal, wang2024adaptivePPO, dickness2025MAPPO}. Adaptations for flood scenarios involve risk-aware rewards or priority queues~\cite{miletic2022review, mic2462025review}, yet face two key issues: (1) convergence relies on high-fidelity simulations, limiting generalization and yielding local optima under sparse feedback~\cite{abdulhai2013MARLIN, zhang2023Safelight}; (2) in high-dimensional environments, policies often collapse due to state explosion~\cite{gao2017experienceReplay}. RL also lacks symbolic reasoning and natural language interfaces, hindering interpretability and collaboration.

\subsection{LLM-Based Strategy Generation Approaches}

LLMs are strong in semantic reasoning and hierarchical planning. Systems like HuggingGPT and multi-agent LLMs demonstrate structured orchestration for smart city tasks~\cite{kalyuzhnaya2025LLMAgents, campo2025WildfireGPT}. However, flood applications expose three core challenges: (1) unstructured outputs lead to semantic drift and hallucinations; (2) semantic–execution gaps arise due to poor alignment with agent control; and (3) lack of feedback prevents adaptation in dynamic settings~\cite{liu2025UrbanLLMAgents}.

\subsection{Multi-Agent Systems and Structural Modeling}

MAS models, including decentralized and GNN-enhanced variants, support spatial task distribution and collaborative routing~\cite{zhao2022GMHM, liu2021hierarchicalAgents}. In flood contexts, agents simulate crowd flow or obstruction propagation, but are typically driven by static rules or RL-based heuristics without semantic grounding or rescheduling~\cite{miletic2022review, mic2462025review}. Graph-based approaches focus on topology but often overlook communication delays and coordination under stress.

\subsection{Modeling Innovation in This Work}

We propose the H–J framework to address these limitations via (i) dual-channel knowledge indexing for semantic grounding; (ii) entropy-constrained hierarchical generation for output stability; and (iii) macro-objective-guided feedback optimization. These modules enable a closed-loop, interpretable, and resilient decision-making system beyond prior paradigms.
Beyond task-specific methods, a complementary line of research frames decision-making as \emph{computational experiments} for \emph{cyber–physical–social systems} (CPSS), systematizing model customization, evaluation, domain-model docking, and integrated experiment systems. This perspective offers controlled yet realistic settings for iterative assessment and generative explanation, which is broadly relevant to urban emergency scheduling~\cite{xue2023AAS, xue2024TSMC, xue2022TCSS_partI, lu2022TCSS_partII, xue2024TCSS_partIII, xue2024JAS_integrated, xue2024JAS_design}. Related CE-based modeling in social manufacturing (SLE/SLE2) also demonstrates experiment-driven, multi-agent coordination at scale~\cite{xue2019TII_SLE, zhou2022TII_SLE2}.

\section{Methodology}

To address challenges in strategy generation, execution instability, and delayed feedback under urban flood dynamics, we propose the H–J framework: a hierarchical multi-agent system driven by entropy constraint ($H$) and a global objective function ($J$), supported by large language models (LLMs). H–J forms a closed-loop architecture of \textit{generation–execution–feedback}, mitigating semantic drift and execution gaps via knowledge grounding and entropy regularization~\cite{liu2025UrbanLLMAgents}. Feedback-driven adjustment enables dynamic adaptation in uncertain environments~\cite{ge2025ufdt}. The framework includes four modules: a strategy pipeline, dual-channel knowledge indexing, entropy-constrained generation, and feedback optimization.

\subsection{Framework Overview and Core Mechanism}
\label{sec:framework-overview}

\begin{figure}[htbp]
\centering
\includegraphics[width=0.7\columnwidth]{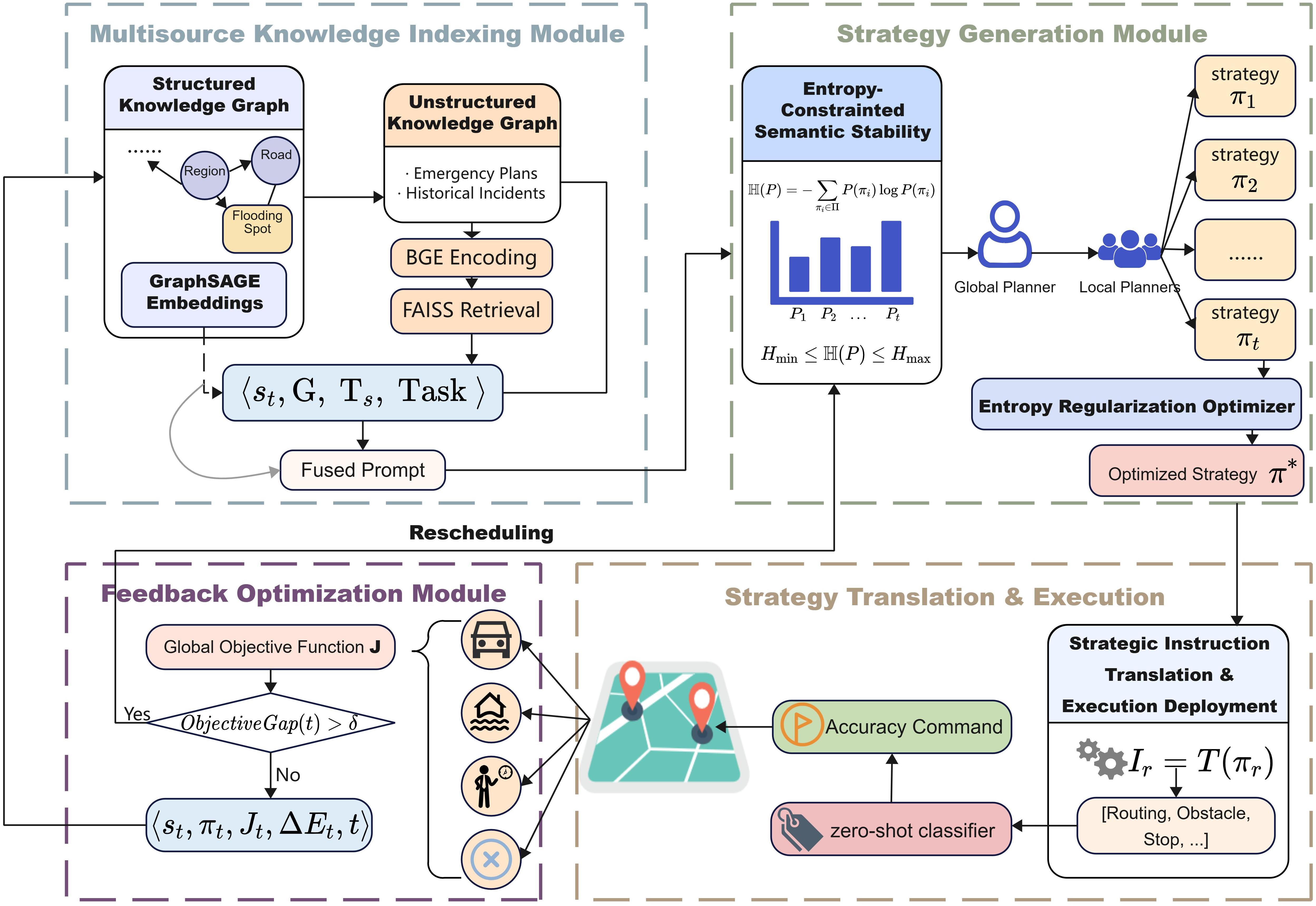}
\caption{
Overview of the H–J framework, which integrates: 
(1) \textbf{Multisource Knowledge Indexing} for structured-unstructured prompt fusion; 
(2) \textbf{Entropy-Constrained Strategy Generation} for robust planning; 
(3) \textbf{Strategy Translation and Execution} for agent-level control; 
(4) \textbf{Feedback Optimization} for adaptive rescheduling based on objective evaluation.
}
\label{fig:tech_pipeline}
\end{figure}

H–J executes in a closed-loop cycle: \textit{retrieval → generation → translation → execution → evaluation → rescheduling}. The LLM parses environment context via dual-channel knowledge retrieval and produces high-level strategies, addressing generalization issues in traditional RL.

Functional agents refine these strategies locally to reduce coordination overhead. Instructions are translated and dispatched in parallel. Outcomes are assessed by the objective function $J$. When deviation exceeds threshold (\textit{ObjectiveGap}), new prompts are generated for replanning.

By fusing symbolic reasoning and adaptive control, H–J improves both decision quality and execution robustness. Its core mechanisms include:

\begin{itemize}
    \item \textbf{Dual-channel knowledge injection}: Combines structured graphs and unstructured text using GraphSAGE and FAISS retrieval to construct grounded prompts.
    \item \textbf{Hierarchical strategy generation}: Employs a Policy–Functional Agent hierarchy with entropy-controlled sampling for balanced stability and diversity.
    \item \textbf{Objective-guided feedback}: Uses structured metrics and adaptive thresholds for continual policy refinement.
\end{itemize}

\begin{algorithm}[h]
\caption{H--J Framework}
\label{alg:hj-framework}
\begin{algorithmic}[1]
\Require Global state $S(t)$, Knowledge base $\mathcal{KB}$, Regional partitions $\mathcal{R} = \{r_1, r_2, \dots, r_n\}$, Execution agents $\mathcal{A}$
\Ensure Updated system feedback $F(t+1)$

\Function{Decision\_Cycle}{$S(t), \mathcal{KB}$}
    \State $\pi_{\text{global}} \gets$ \Call{LLM\_Policy\_Generate}{$S(t), \mathcal{KB}$}
    \ForAll{$r \in \mathcal{R}$}
    \State $\pi_r \gets$ \Call{RoleAgent\_Generate\_Local}{
    $\pi_{\text{global}},\ S_r(t)$%
    }
    \State $\text{Instr}_r \gets$ \Call{Translate}{$\pi_r$}
    \State \Call{Dispatch}{$\text{Instr}_r, \text{Agent}_r$}
    \EndFor
    \ForAll{$a \in \mathcal{A}$}
        \State $a.\text{execute}()$
        \State $e \gets a.\text{observe\_local\_environment}()$
        \State $f \gets a.\text{report\_feedback}(e)$
        \State $F \gets$ \Call{Aggregate}{$F, f$}
    \EndFor
    \If{\Call{ObjectiveGap}{$F, J$} $\geq \delta$}
        \State \Call{Trigger\_Replanning}{$F, \mathcal{KB}$}
    \EndIf
    \State \Return $F$
\EndFunction
\end{algorithmic}
\end{algorithm}
Our retrieval–generation–translation–execution–evaluation loop echoes the principles of \emph{domain-model docking} and \emph{integrated experiment-system design} advocated by the computational-experiment community, where high-level strategies are grounded into executable components inside simulators and operations~\cite{xue2024TCSS_partIII, xue2024JAS_integrated}. This lens also supports systematic what-if testing and \emph{generative explanation} for decision outcomes~\cite{xue2024JAS_design}.
\subsection{Knowledge Indexing: Multisource Fusion and Semantic Control}
\label{sec:knowledge-indexing}

In urban flood scenarios, LLMs often suffer from unstable semantic grounding. Standard prompt construction or retrieval-augmented generation (RAG) techniques struggle to align heterogeneous knowledge, leading to hallucinations and poor instruction fidelity~\cite{yu2022heteroRAG}.

To address this, we design a dual-channel knowledge indexing module that fuses structured and unstructured sources into task-specific prompts. This aligns with recent dual-path RAG architectures that retrieve from both knowledge graphs and textual corpora, then merge them into coherent contexts~\cite{chen2024KGRetriever, wu2025KGInfusedRAG}.

\paragraph{Structured Knowledge Channel.} We use \textbf{GraphSAGE} to encode the urban flood knowledge graph $G=(V,E)$, where nodes represent regions, roads, and flood spots. Multi-hop aggregation yields structural embeddings resilient to sparsity and adaptable to dynamic conditions.

\paragraph{Unstructured Knowledge Channel.} A pretrained \textbf{BGE-Encoder} transforms emergency reports and historical logs into semantic vectors. This enhances retrieval of relevant unstructured content based on system context $s_t$.

\paragraph{Context-Aware Retrieval.} The current state $s_t$ is encoded into a query vector $q_t = \text{Encoder}(s_t)$. For structured knowledge, a task-relevant subgraph $G_s = \text{SubgraphExtract}(G_t, s_t)$ is extracted. For unstructured data, similarity scores $\text{Sim}(q_t, K_t)$ guide top-$K$ segment retrieval, ensuring semantic alignment.

\paragraph{Prompt Construction.} Retrieved content is fused into a hybrid prompt:
\begin{equation}
P_t = \langle s_t, G, T_s, \text{Task} \rangle
\end{equation}
This design bridges symbolic and semantic representations while enabling prompt adaptation to environmental changes.

\paragraph{Graph Update.} The knowledge graph evolves via:
\begin{equation}
G(t+1) = G(t) \cup \{v_{\text{new}}, e_{\text{new}}\}
\end{equation}
New entities or relations are integrated during generation, improving knowledge freshness and structural consistency.

In summary, the dual-channel indexing mechanism grounds LLM prompts in both symbolic and semantic spaces, supporting coherent, adaptive decision-making under high-stakes, dynamic urban conditions.

\subsection{Hierarchical Strategy Generation: Semantic Decomposition and Entropy-Constrained Stabilization}

Although large language models (LLMs) are capable of complex strategy generation, they often exhibit semantic instability and inconsistent execution in urban flood scenarios. Traditional sampling methods (e.g., temperature scaling or top‑$k$ truncation) introduce stochasticity that amplifies semantic drift under ambiguous inputs, making them insufficient for stable, controllable outputs~\cite{foodeei2025semanticUncertainty, minh2025minPSampling, verine2025precRecall}.

To address this, we propose a hierarchical strategy generation mechanism that combines semantic decomposition with entropy‑constrained policy stabilization. Hierarchical reasoning and task decomposition have been shown to reduce semantic instability and enhance interpretability~\cite{gui2025hyperTree}, while entropy minimization techniques improve decoding consistency and output confidence~\cite{foodeei2025semanticUncertainty}.

\subsubsection{Global-to-Local Strategy Hierarchy}

The generation process is structured hierarchically: a global-level \textit{Policy Agent} issues macro strategies based on system-wide states, while local agents refine these into fine-grained actions. Let $s$ be the global observation and $\mathcal{A}_h$ the high-level action space. The entropy of the global strategy $\pi_h$ is defined as:
\begin{equation}
H(\pi_h) = - \sum_{a \in \mathcal{A}_h} \rho_\theta(a|s) \log \rho_\theta(a|s)
\end{equation}
where $\rho_\theta(a|s)$ is the probability of high-level action $a$ under state $s$.

To enforce semantic consistency across levels, we introduce a conditional entropy constraint ensuring that local policies $\pi_l$ do not exceed global uncertainty:
\begin{equation}
H(\pi_l \mid \pi_h) = \mathbb{E}_{a_h \sim \pi_h} \left[ H\left(\pi_l(\cdot \mid a_h)\right) \right]
\end{equation}
\begin{equation}
H(\pi_l \mid \pi_h) \leq H(\pi_h), \quad \text{and} \quad H(\pi_h) \leq \tau
\end{equation}
Here, $\tau$ is a global entropy threshold that limits uncertainty propagation, supporting structured semantic alignment.

\subsubsection{Explicit Entropy-Constrained Loss}

To incorporate this constraint into training, we define an entropy-regularized objective:
\begin{equation}
\mathcal{L} = \mathbb{E}_{\pi_\theta} \left[ \log \rho_\theta(\pi) \right] - \lambda \cdot \left| H(\rho_\theta) - \tau \right|
\end{equation}
This loss balances log-likelihood maximization and entropy suppression, stabilizing policy outputs by penalizing excess uncertainty. The coefficient $\lambda$ dynamically adjusts the penalty strength.

\subsubsection{Multi-Level Fine-Tuning and Inference Flow}

Training is staged to maintain coherence across strategy levels:
\begin{itemize}
    \item \textbf{High-level tuning:} The \textit{Policy Agent} is trained under entropy constraints to align global semantics.
    \item \textbf{Local-level tuning:} Regional agents refine global strategies using local observations to produce executable commands.
    \item \textbf{Entropy coefficient update:} To avoid under- or over-regularization, $\lambda$ is updated by:
    \begin{equation}
    \lambda_{t+1} = \lambda_t + \alpha \cdot \left( H(\pi_h) - \tau \right)
    \end{equation}
    where $\alpha$ is a tunable rate controlling convergence.
\end{itemize}
This hierarchical process supports flexible expansion or contraction of the policy space, enhancing robustness under extreme flood conditions and preserving generalization.

\subsection{Closed‑Loop Feedback Optimization: Dynamic Strategy Substitution Guided by Macro‑Objective Functions}

In fast‑changing flood scenarios with spatial heterogeneity and behavioral volatility, fixed‑strategy baselines often lead to unstable outcomes. To enable real‑time adaptability, we embed a closed‑loop feedback mechanism driven by the global objective function $J$, allowing the model to refine policies based on deviations between actual and expected outcomes~\cite{kim2024RLMPCtraffic, ge2025ufdt}.
This design extends adaptive control and Model Predictive Control (MPC) principles to dynamic policy substitution, supporting real‑time strategy adjustment under stochastic environmental feedback~\cite{gomes2024detentionPonds, feher2025pathplanRL}.
\subsubsection{Weighted Macro-Objective Design}

While Pareto-based optimization offers theoretical rigor, it lacks scalar interpretability for real-time control and incurs high computational costs in evaluating tradeoffs. Thus, we design a weighted linear fusion of core objectives into a single evaluation function:
\begin{equation}
J = \omega_1 f + \omega_2 t + \omega_3 c + \omega_4 (1 - r)
\end{equation}
Here, $f$ (flood risk), $t$ (traffic congestion), $c$ (trip cancellation), and $r$ (arrival rate) are normalized indicators. The weights $\omega_i$ are adapted via experience replay and short-term performance tracking to reflect shifting priorities in evolving urban contexts. Detailed formulations of the normalized indicators $f$ and $t$ are provided in the supplementary material.

\subsubsection{Deviation-Based Feedback Triggering}

To detect policy degradation, we compute performance gaps relative to historical bests:
\begin{equation}
\text{ObjectiveGap}(t) = J(t) - \min_{\tau < t} J(\tau)
\end{equation}
Instead of static thresholds, we apply an adaptive feedback trigger:
\begin{equation}
\delta = \mu_J + \lambda \sigma_J
\end{equation}
where $\mu_J$ and $\sigma_J$ are the mean and standard deviation over recent $J$ values. This ensures stable yet responsive feedback control—suppressing noise while activating meaningful corrections.

Upon trigger activation, prompt regeneration incorporates symbolic failure feedback (e.g., intention mismatches, local execution gaps), supporting closed-loop semantic evolution. To measure deviation severity:
\begin{equation}
\Delta E = \sqrt{ \frac{1}{|X|} \sum_{x \in X} (x^{\text{exec}} - x^{\text{plan}})^2 }
\end{equation}
where $x^{\text{exec}}$ and $x^{\text{plan}}$ are executed and planned values for each metric $x$ (e.g., trip arrival, congestion). These feedback signals are embedded into the knowledge graph for future reference.

In summary, this closed-loop feedback integrates symbolic updates, numerical deviation tracking, and prompt rewriting. It supports long-horizon adaptation and serves as the foundation for real-time large-model strategy evolution in urban flood resilience.

\section{Experimental}
\begin{table*}[h]
\centering
\renewcommand{\arraystretch}{1.2}
\caption{Model and System Configuration of the H--J Framework}
\label{tab:hj-config}
\begin{tabular}{@{}l l l@{}}
\toprule
\textbf{Category} & \textbf{Parameter} & \textbf{Value} \\
\midrule
\multirow{4}{*}{LLM}
& High-level model & Qwen-7B-Chat (LoRA + Entropy Loss) \\
& Regional policy model & GLM-6B (LoRA) \\
& Optimizer & AdamW ($\beta_1{=}0.9$, $\beta_2{=}0.98$) \\
& Learning rate / Batch size & 3e-5 / 32 \\
\midrule
\multirow{3}{*}{Entropy Control}
& Entropy threshold $\tau$ & 1.2 \\
& Penalty coefficient $\lambda$ & Init.\ 1.0 (dynamic) \\
& Rhythm modulation rate $\alpha$ & 0.05 \\
\midrule
\multirow{3}{*}{Multi-Agent}
& Region partitions & 64 \\
& Initial agent population & 500 \\
& Perception radius & 3 \\
\midrule
\multirow{3}{*}{Feedback Mechanism}
& Weight vector $\bm{w}$ & $[0.3, 0.3, 0.2, 0.2]$ \\
& Trigger threshold $\delta$ & 0.015 \\
& Feedback window length & 10 \\
\bottomrule
\end{tabular}
\vspace{0.4em}

\parbox{\textwidth}{\small\textit{\textbf{Note:}
Experiments were run under three rainfall scenarios, each strategy repeated $\geq 5$ times.
We recorded the mean/variance of key metrics, agent trajectories, and semantic outputs for trend analysis.
Metric $J$ and semantic indices were tracked across iterations. All hyperparameters were tuned via preliminary runs.
Experiments were conducted on dual RTX 4090 GPUs (24GB), using LoRA with 4-bit quantization for entropy-regularized fine-tuning.}}
\end{table*}

To evaluate the performance and theoretical strengths of the proposed H--J framework, we conduct simulation experiments under representative urban flood scenarios. The evaluation focuses on three aspects: overall strategy effectiveness, module-wise contribution, and system robustness.

We investigate the following research questions:

\begin{itemize}
    \item \textbf{Q1: System-Wide Effectiveness.} Can H--J significantly improve dispatch quality under dynamic urban flooding? This is assessed via indicators such as traffic mitigation and task completion.

    \item \textbf{Q2: Module-Wise Contribution.} Do the three core mechanisms—dual-channel knowledge retrieval, entropy-constrained generation, and objective-guided feedback—offer measurable and complementary benefits? Can their marginal contributions to the global objective be quantitatively verified?
\end{itemize}

To address these questions, we conduct multi-round comparative experiments, ablation studies for each mechanism, and apply structured performance metrics along with semantic-level evaluations. This comprehensive design enables rigorous validation of H--J’s technical feasibility and theoretical innovation in adaptive urban resilience modeling.

\subsection{Experimental Setup}
\subsubsection{Scenario Construction and Environmental Settings}

To validate the adaptability and coordination of the H–J framework, we construct a realistic multi-source environment integrating rainfall, mobility, and transit dynamics in flood-prone cities.

\paragraph{Environmental Data Sources}
We incorporate (1) regional rainfall and waterlogging data from a municipal meteorological bureau in East Asia (via an affiliated water management institute),(2) Hexi District bus networks with dynamic rerouting, and (3) resident mobility modeled via POIs and A*-based pathfinding over grid networks. The simulation is built on Mesa ABM with real-time environment-agent feedback.

\paragraph{Agent State Modeling (see supplementary material)}
Each agent operates under a structured state-action framework to support adaptive decision-making in dynamic environments.

\subsubsection{Model Configuration and System Parameters}

To ensure reproducibility and scalability, all modules are integrated into a unified Mesa + HuggingFace platform. Configuration details are shown in Table~\ref{tab:hj-config}.

\subsubsection{Evaluation Metrics and Methods}

We adopt two categories of evaluation metrics:

\paragraph{A. Macroscopic Execution Metrics (for \(J\))}

The unified objective \(J\) integrates four normalized indicators: average water depth (\(f\)) for inundation risk, average travel time (\(t\)) for traffic congestion, failed task ratio (\(c\)) as cancellation rate, and successful arrival ratio (\(r\)) for arrival effectiveness.

\paragraph{B. Semantic-Level Evaluation}

We further assess LLM output robustness through:

\begin{itemize}
  \item \textbf{Semantic Consistency Score (SCS)}: average similarity across same-prompt responses; semantic consistency has been rigorously defined and evaluated in QA and generative settings using embedding similarity or entailment measures~\cite{rabinovich2023semanticConsistency, nalbandyan2025SCORE}.
  \item \textbf{Semantic Diversity Score (SDS)}: embedding spread across agents under shared input; similar to metrics used in domain generalization research to measure output diversity while preserving semantic content~\cite{niu2024SCSD}.
  \item \textbf{Execution Stability}: variance of \(f, t, c, r\) across runs.
\end{itemize}
The computation methods for SCS and SDS are detailed in the supplementary material.

\subsection{Comparison Experiment Design}

We benchmark H--J against four baselines:

\begin{itemize}
    \item \textbf{Empty:} No planning or coordination; agents cancel when blocked.
    \item \textbf{Ruled:} Static heuristic rules without environmental awareness.
    \item \textbf{PPO:} Reinforcement learning with PPO, no external knowledge or entropy control.
    \item \textbf{H--J:} Full framework integrating dual-indexing, entropy constraints, and feedback optimization.
\end{itemize}

All strategies run under identical initialization across three rainfall scenarios, each repeated five times with mean and variance reported.

\subsection{Ablation Experiment Design}

To assess the marginal impact of key mechanisms, we design three ablation variants:

\begin{itemize}
    \item \textbf{w/o Dual Indexing:} Removes structured + semantic retrieval guidance.
    \item \textbf{w/o Entropy Control:} Disables entropy-constrained generation.
    \item \textbf{w/o Feedback Loop:} Disables dynamic objective-guided replanning.
\end{itemize}

All variants preserve the full architecture except for the targeted component, ensuring fair comparison.

\subsection{Overall Performance Validation (Q1)}

\begin{figure}[h]
  \centering
  \subfloat{\includegraphics[width=0.7\linewidth]{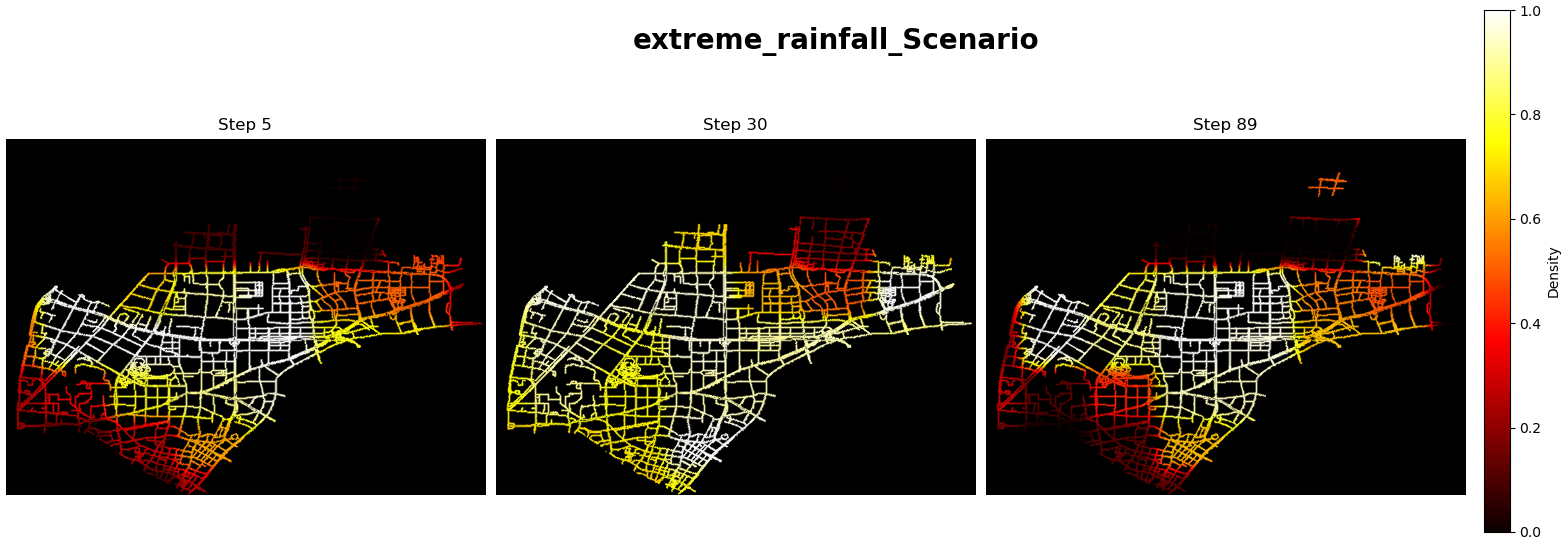}} \\
  \subfloat{\includegraphics[width=0.7\linewidth]{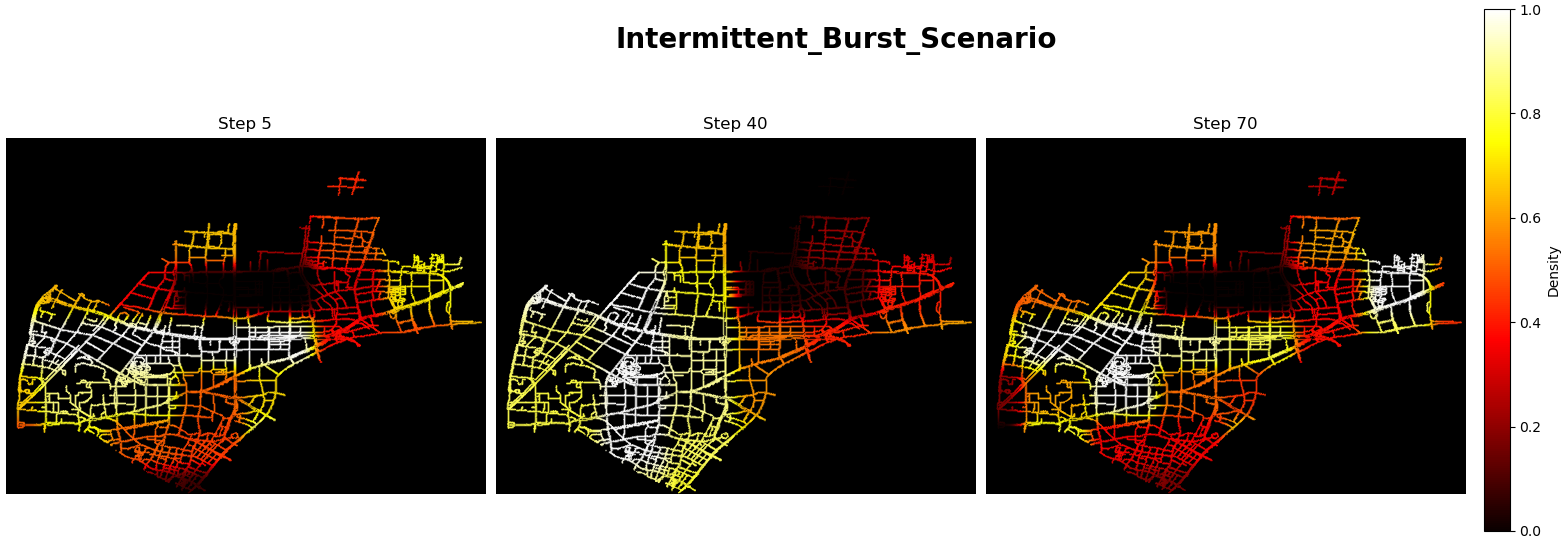}} \\
  \subfloat{\includegraphics[width=0.7\linewidth]{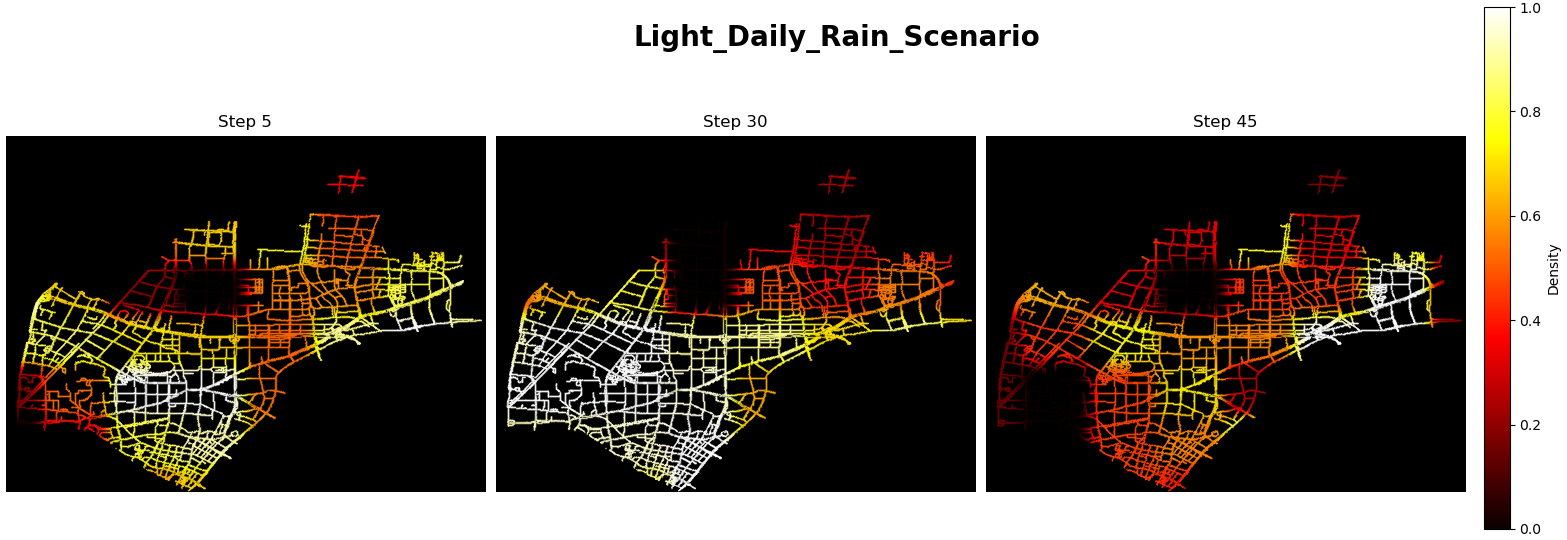}}
  \caption{Human flow density heatmaps at representative steps (5, 30/40, 45). The H--J strategy dynamically shifts flows away from high-risk or congested regions. In the intermittent burst case (middle row), load in the northern trunk road is significantly relieved by step 40.}
  \label{fig:human_flow_heatmap}
\end{figure}
To answer \textbf{Q1}—whether the H–J framework improves dispatch effectiveness—we conduct a three-level analysis: 
(1) trend analysis of normalized inundation levels; and 
(2) human flow heatmap visualization across three rainfall scenarios.
(3) comparative evaluation across four representative strategies (H–J, PPO, rule-based, and no-policy baseline); 

These complementary analyses collectively assess both outcome-oriented metrics (e.g., objective $J$, inundation stability) and behavioral patterns under varying external pressures. Although inundation patterns are exogenous, they form a dynamic backdrop to test strategy robustness. H–J must adapt to this evolving pressure to maintain operational continuity.

\subsubsection{Comparative Results Across Strategies}
\begin{figure}[t]
  \centering
  \subfloat{%
    \includegraphics[width=.5\linewidth,keepaspectratio]{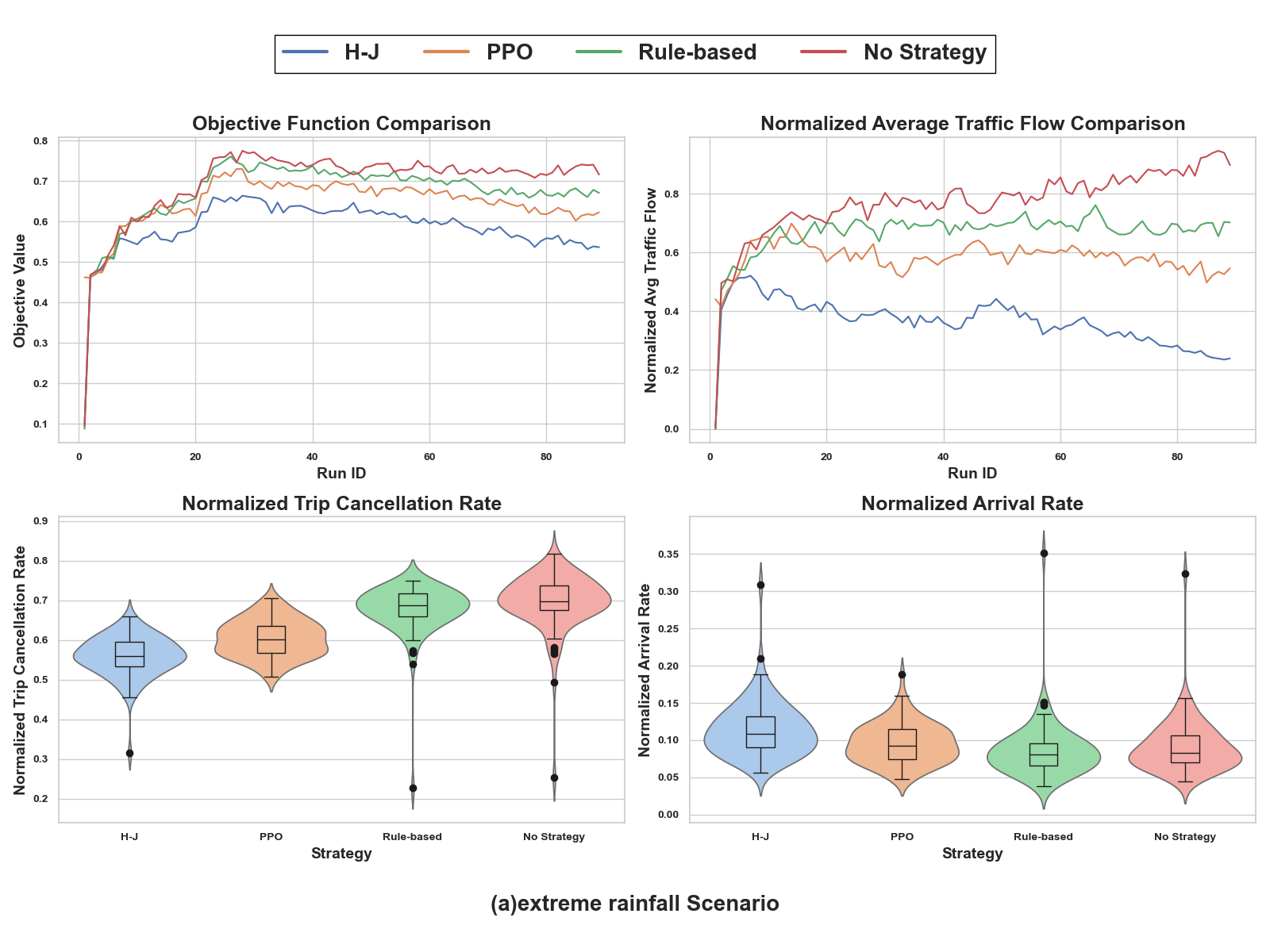}}%
  \par\vspace{2pt} 

  \subfloat{%
    \includegraphics[width=.5\linewidth,keepaspectratio]{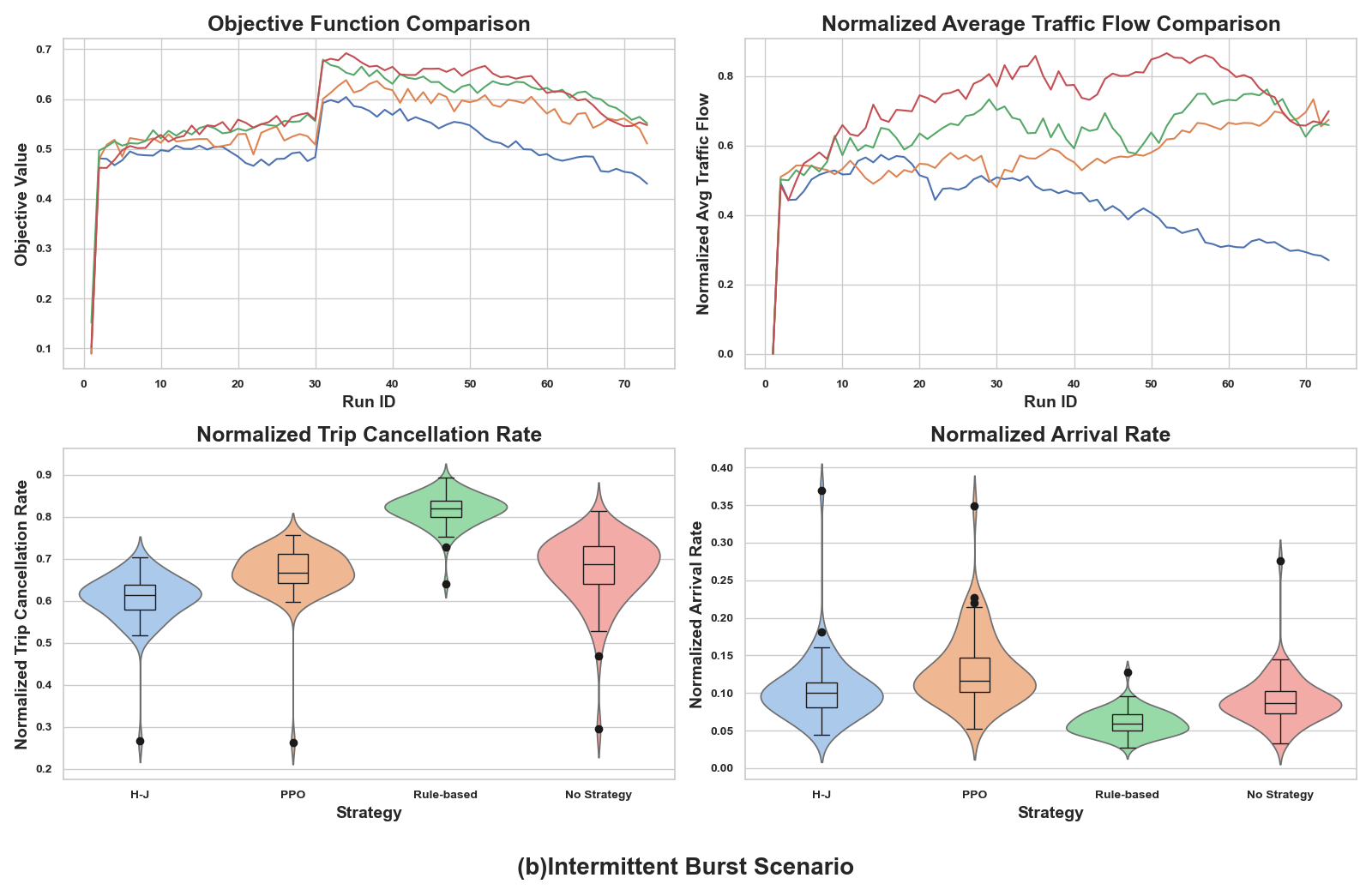}}%
  \hfill
  \subfloat{%
    \includegraphics[width=.5\linewidth,keepaspectratio]{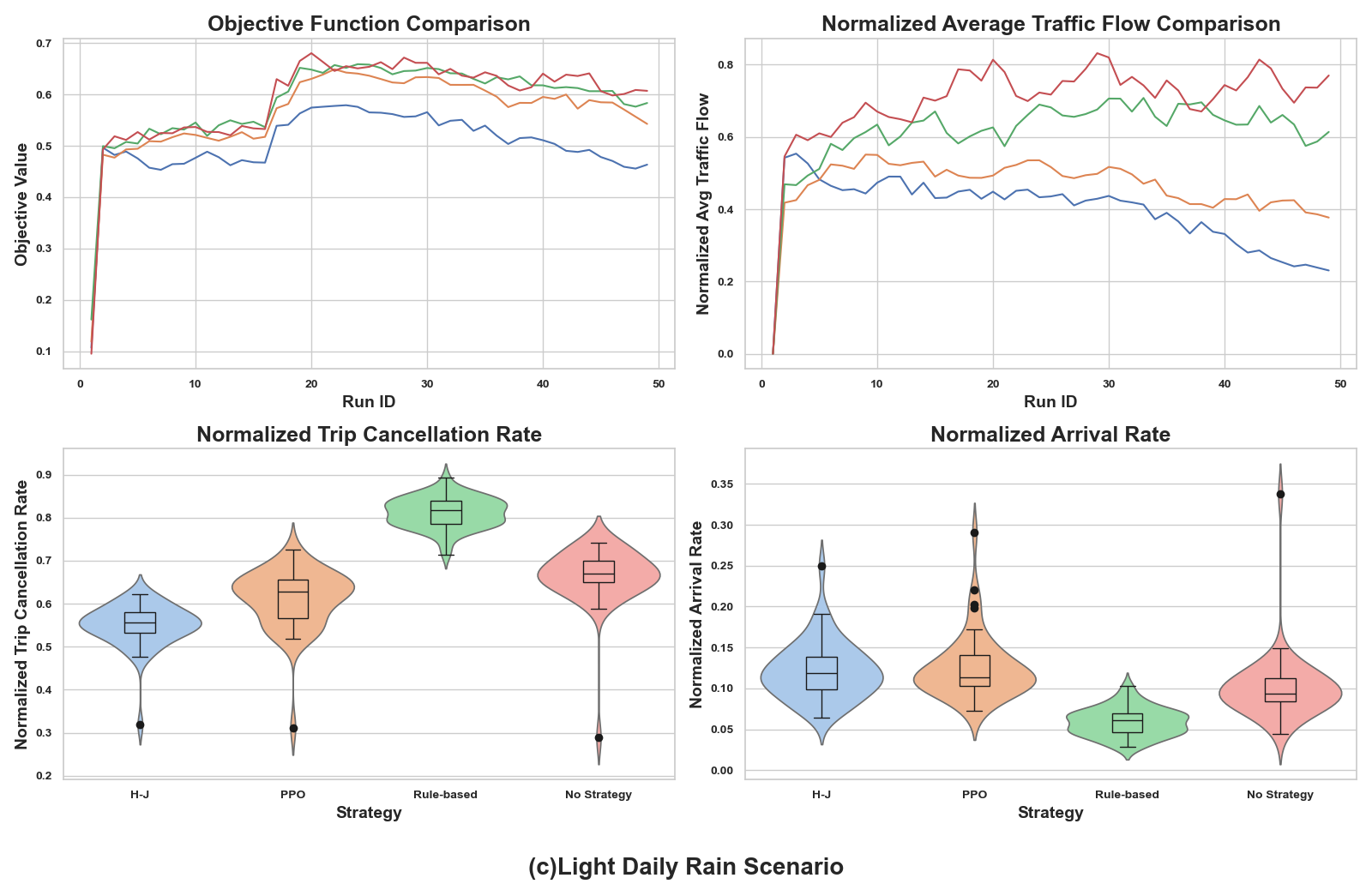}}%

  \caption{Comparative performance under three rainfall scenarios. H--J consistently outperforms baselines across key indicators.}
  \label{fig:strategy_comparison}
\end{figure}

We compare four strategies—Empty, Rule-based, PPO, and H--J—under identical conditions using four indicators: objective $J$, traffic flow, cancellation rate, and arrival rate.

H--J exhibits the most stable $J$ trajectory, indicating strong coordination and feedback regulation. PPO converges quickly but fluctuates more under environmental shifts. Rule-based and Empty strategies lack adaptability and fail to achieve operational efficiency.

In the \textit{intermittent burst} scenario, PPO slightly surpasses H--J in arrival rate, suggesting a tendency toward short-term gains at the expense of system-wide stability. Conversely, H--J optimizes a global objective and exhibits conservative, risk-aware behavior. H--J achieves its best results in the \textit{light daily rain} setting and remains robust even under \textit{extreme} rainfall.

In summary, the H--J framework balances strategic foresight and local adaptability, delivering superior outcomes across a range of flood intensities.

\vspace{0.5em}
\noindent\textit{Details on simulation dynamics, risk mitigation patterns, and indicator derivation are provided in supplementary material.}

\subsection{Ablation Study: Module-wise Contribution (Q2)}

\begin{figure}[htbp]
  \centering
  \includegraphics[width=0.95\linewidth]{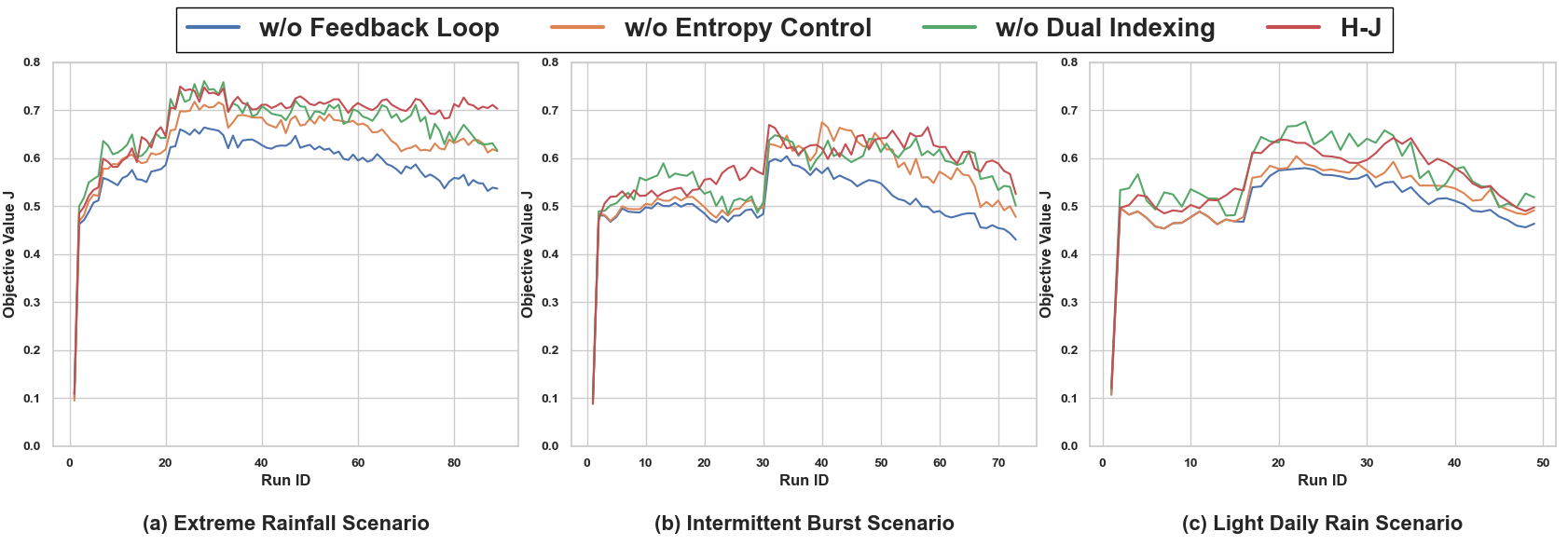}
  \caption{Ablation results under three rainfall scenarios. Removing any module causes performance drops, especially under challenging conditions, confirming the necessity of all three components.}
  \label{fig:ablation_q2}
\end{figure}

To address \textbf{Q2}—the contribution of H--J’s three core mechanisms (dual-channel knowledge indexing, entropy-constrained control, and feedback optimization)—we conduct ablation experiments under three rainfall scenarios, disabling one module at a time while keeping others intact (see Fig.~\ref{fig:ablation_q2}).

The full H--J system consistently achieves the highest objective value $J$. The most pronounced decline occurs when the \textit{dual-channel knowledge indexing} is removed, especially under extreme rainfall, highlighting the importance of structured–unstructured knowledge fusion. Omitting the \textit{entropy control} leads to higher volatility, while removing the \textit{feedback mechanism} weakens long-term adaptability.

\subsubsection{Quantitative Contribution of Core Modules}

\begin{table}[htbp]
  \centering
  \caption{Quantitative ablation results. The full H--J framework achieves the best output stability, semantic consistency (SCS), and strategy diversity (SDS).}
  \label{tab:module_metrics}
  \begin{tabular}{lccc}
    \toprule
    \textbf{Module Setting} & \textbf{Stability $\downarrow$} & \textbf{SCS $\uparrow$} & \textbf{SDS $\uparrow$} \\
    \midrule
    H--J Full Framework      & 0.0047 & 0.872 & 0.443 \\
    w/o Feedback             & 0.0059 & 0.858 & 0.397 \\
    w/o Entropy Control      & 0.0061 & 0.851 & 0.312 \\
    w/o Dual Indexing        & 0.0071 & 0.762 & 0.421 \\
    \bottomrule
  \end{tabular}
\end{table}

As Table~\ref{tab:module_metrics} shows, the full model exhibits the lowest variance and the highest semantic scores. Notably, entropy control raises SDS from 0.312 to 0.443 and improves SCS by 0.021, revealing its role in balancing diversity and coherence. Dual indexing boosts SCS by over 0.11 (from 0.762 to 0.872), ensuring semantic precision through structured semantic grounding. Meanwhile, feedback reduces variance by 20\%, stabilizing long-horizon strategy adaptation.

Together, the three modules enhance the framework's semantic accuracy (indexing), generation diversity and robustness (entropy), and dynamic responsiveness (feedback). Their complementary contributions form the backbone of the H--J architecture’s robustness under complex flood dynamics.
\section{Conclusion}

This paper presents the H--J framework, a hierarchical large-model-driven approach for urban flood dispatching. It integrates dual-channel knowledge indexing, entropy-constrained guidance, and a macro-objective-based feedback optimization mechanism. The proposed method enables multi-agent coordination under different roles to achieve global policy control, significantly improving traffic efficiency and task completion rates under various rainfall scenarios.

Experimental results demonstrate that the H--J framework consistently outperforms traditional rule-based and mainstream reinforcement learning approaches in terms of semantic consistency, strategy diversity, and execution stability. The ablation analysis further verifies the synergistic contributions of the three core modules, particularly highlighting the dual role of knowledge indexing in improving semantic alignment and response diversity, and the entropy constraint’s role in stabilizing generation and enhancing diversity tradeoffs.

Despite its superior empirical performance, the H--J framework still faces practical limitations, such as high deployment cost and reliance on large-scale inference resources. Future work will explore lighter-weight coordination schemes and more efficient modular knowledge fusion mechanisms, with the aim of enhancing their deployability and real-world applicability.
A promising direction is to further systematize our pipeline under a \emph{computational-experiment/CPSS} paradigm to strengthen explainability, transferability, and evaluation fidelity across urban risks~\cite{xue2024TSMC, xue2023AAS}.
\bibliographystyle{unsrtnat}  
\bibliography{references}     
\appendix
\appendix
\setcounter{secnumdepth}{3} 
\renewcommand{\thesection}{\Alph{section}}
\renewcommand{\thesubsection}{\Alph{section}.\arabic{subsection}}
\renewcommand{\thesubsubsection}{\Alph{section}.\arabic{subsection}.\arabic{subsubsection}}

\section{Simulation Environment Details}

To evaluate the adaptability of the H--J framework under varying flood risks and environmental dynamics, we simulate three representative rainfall scenarios: \textit{extreme rainfall}, \textit{intermittent bursts}, and \textit{daily light rain}. These scenarios capture diverse patterns of urban waterlogging stress and are constructed based on synthesized rainfall intensity curves projected over time.

\subsection{Rainfall Scenario Design}

\begin{itemize}
    \item \textbf{Extreme Rainfall:} Characterized by a sharp onset and sustained high intensity over a short period. This scenario induces rapid inundation and severely disrupts agent mobility, testing the system's emergency coordination capacity.
    
    \item \textbf{Intermittent Bursts:} Composed of several high-intensity pulses with dry intervals in between. This setting challenges the system's reactivity and strategy switching ability under fluctuating pressures.
    
    \item \textbf{Light Daily Rain:} A low-intensity but long-duration scenario, simulating typical seasonal rainfall. It tests whether the system can maintain stable performance over prolonged but manageable stress.
\end{itemize}

All rainfall curves are discretized into simulation steps, and mapped to spatial grids via a runoff-constrained water diffusion model. The resulting water depth evolution is dynamically updated at each step and fed into the agent perception module as part of $S_t$.

\subsection{Rainfall Trend Visualization}

\begin{figure}[htbp]
    \centering
    \includegraphics[width=0.48\textwidth]{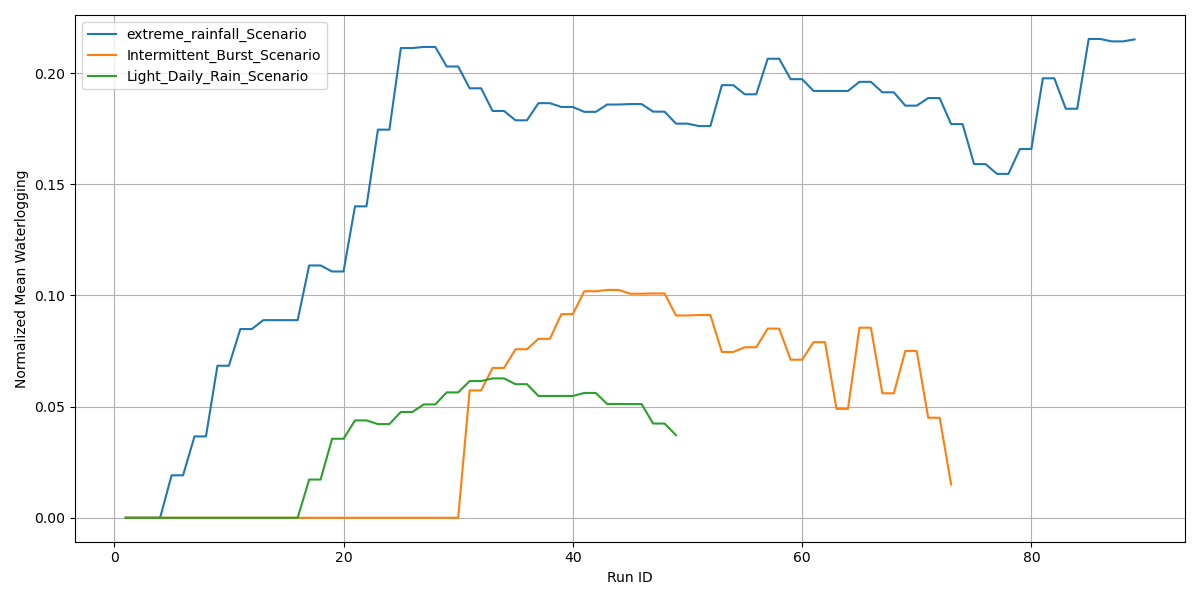}
    \caption{Normalized rainfall intensity trends for extreme, intermittent, and light rain scenarios.}
    \label{fig:rainfall-trend}
\end{figure}

Figure~\ref{fig:rainfall-trend} illustrates the temporal evolution of normalized rainfall intensity across the three scenarios. Each curve is scaled to a common time axis and normalized to $[0,1]$ for comparability.

\section{Agent State Design}

To enable context-aware and adaptive strategy execution, each agent in the H--J framework maintains a structured, perception-driven state representation. The system simulates a dynamic multi-agent environment with diverse roles including transit agents and resident agents, all interacting over a high-fidelity grid-based urban simulation.

\subsection{Environmental Data Sources and Integration}

The environment integrates three categories of data:

\begin{itemize}
    \item \textbf{Rainfall and inundation data:} Synthetic rainfall patterns are generated from historical meteorological statistics and mapped to grid cells using a topography-aware waterlogging simulator. The resulting flood depths dynamically evolve and serve as a key environmental input.

    \item \textbf{Public transit data:} Bus lines and station locations define the routing and behavior of \texttt{BusAgent}s, who adaptively re-plan based on road inundation conditions.

    \item \textbf{Resident mobility data:} \texttt{ResAgent}s generate travel demands based on sampled POI distributions and execute routes using A*-based planning with stochastic detours. All agent movements are projected onto a unified spatial network and aggregated into semantic flow indicators (e.g., trip density, route congestion).
\end{itemize}

These multi-source inputs are unified in real-time to simulate an urban dispatch scenario under evolving flood stress.

\subsection{Perception-Driven State Formulation}

Each agent maintains an internal state to support proactive sensing and decision-making. At each time step, the agent interacts with its environment, receives multimodal observations, and selects actions accordingly. The environment state is formalized as a 4-tuple:

\[
(S_t, \mathcal{A}_t, \delta, \pi)
\]

where:

\begin{itemize}
    \item $S_t$: Current environmental observation, including grid-wise rainfall intensity, water depth, and local traffic status.
    \item $\mathcal{A}_t$: Available action set at time $t$, such as route choices, detour options, or cancellation decisions.
    \item $\delta$: Environment transition function, encoding how the system evolves in response to agent actions and exogenous dynamics (e.g., rainfall, congestion propagation).
    \item $\pi$: A probabilistic policy over $\mathcal{A}_t$, representing the agent's uncertainty-aware decision distribution based on partial observability.
\end{itemize}

This state abstraction supports both short-term reactive behaviors and long-term adaptive planning. Agents continually update their internal state using recent feedback and environmental sensing, enabling stable policy execution under real-world uncertainties.

\section{Strategy Translation and Execution Wrapping}

To ensure the executable reliability of language-based strategies, the H--J framework integrates a structured translation and execution module that converts regional natural language strategies into system-compatible action representations. This enables seamless interfacing between the LLM-generated plans and the downstream control layer of intelligent agents.

\subsection{Translation Function}

Given a regional strategy $\pi_r$ generated by the Local Specialist, the system maps it to a structured command representation via:

\[
I_r = T(\pi_r)
\]

Here:
\begin{itemize}
    \item $\pi_r$: The regional-level natural language strategy.
    \item $T(\cdot)$: The instruction translation function, responsible for parsing, grounding, and packaging the semantic plan into a set of executable action commands.
\end{itemize}

\subsection{Instruction Classification and Mapping}

To enhance task alignment and cross-domain transferability, the system embeds a zero-shot command classifier. It leverages the LLM’s latent generalization ability to assign semantic tags (e.g., \texttt{Routing}, \texttt{Obstacle}, \texttt{Stop}, etc.) to unstructured segments of $\pi_r$, which are then mapped to agent-understandable instructions via a predefined action library.

\subsection{Execution Alignment and Refinement}

To ensure the spatial and temporal alignment of translated commands with the evolving environment, an \textit{Accuracy Command Wrapper} evaluates each translated instruction against local context, road status, and feasibility constraints. The module adjusts parameters such as location anchors, execution windows, and tolerance thresholds to ensure that the final instruction set is both spatially deployable and semantically faithful to the original strategy intent.

\section{Expanded Metric Definitions}

To support reproducibility and provide clarity on evaluation criteria, we detail here the full formulations of the metrics used in the macro-objective function $J$, along with semantic-level evaluation indices.

\subsection{Macro-Objective Function}

The overall evaluation function $J$ is a weighted linear combination of four normalized objectives:

\[
J = \omega_1 f + \omega_2 t + \omega_3 c + \omega_4 (1 - r)
\]

where:
\begin{itemize}
    \item $f$: normalized flood risk index,
    \item $t$: normalized traffic congestion index,
    \item $c$: average trip cancellation rate,
    \item $r$: average trip success rate.
\end{itemize}

\subsection{Design Rationale for Fixed Weights \texorpdfstring{$\omega_i$}{ω}}

In this work, we adopt a fixed set of weights $\omega_i$ in the macro-objective function $J$. While dynamic weight adjustment can offer adaptability, it also introduces instability and additional computational overhead in real-time control. Given the well-defined priority structure of our urban dispatching task---where flood risk and mobility reliability consistently outweigh traffic fluctuations---we found that a carefully tuned fixed weight setting strikes a better balance between interpretability, robustness, and deployment feasibility.

The chosen weights $(\omega_1, \omega_2, \omega_3, \omega_4) = (\alpha, \beta, \gamma, \delta)$ were determined through grid search in preliminary simulations and remain consistent across all experiments. Sensitivity analysis confirmed the stability of relative performance across a range of nearby values. Future work may explore adaptive weighting schemes using Pareto front approximation or multi-objective RL. However, in our current setting, fixed weights provide a practical and stable baseline.

\subsection{Flood Risk Index $f$}

The indicator $f$ measures average inundation severity across regions via sigmoid normalization:

\[
f_a = \frac{1}{1 + \exp\left( -\frac{D_a - \mu_D}{\sigma_D} \right)}, \quad
f = \frac{1}{|\mathcal{A}|} \sum_{a \in \mathcal{A}} f_a
\]

Here, $a \in \mathcal{A}$ denotes a discretized urban sub-region in the simulation grid, and $|\mathcal{A}|$ is the total number of such regions. $D_a$ is the observed water depth in region $a$, while $\mu_D$ and $\sigma_D$ denote the global mean and standard deviation computed over all regions $\mathcal{A}$.

\subsection{Traffic Congestion Index $t$}

The traffic congestion index quantifies regional traffic load based on vehicle density $C_a$ in each sub-region:

\[
t_a = \frac{1}{1 + \exp\left( -\frac{C_a - \mu_C}{\sigma_C} \right)}, \quad
t = \frac{1}{|\mathcal{A}|} \sum_{a \in \mathcal{A}} t_a
\]

Here, $C_a$ denotes the instantaneous car density in region $a$, computed as the number of active vehicles per unit area (vehicles/grid cell) during the simulation. The global mean $\mu_C$ and standard deviation $\sigma_C$ are aggregated across all regions $\mathcal{A}$ over the evaluation period.

\subsection{Trip Cancellation and Arrival Metrics}

\begin{itemize}
    \item $c$: Proportion of canceled trips among all residents due to extreme delay or blockage,
    \item $r$: Proportion of successful trips that reached destinations within a tolerable delay threshold.
\end{itemize}

Note that $r$ is used in the form $(1 - r)$ within the objective function to reflect a penalty term---i.e., lower success rates increase $J$. This transformation ensures alignment of all components in $J$ as cost-oriented indicators.

\subsection{Semantic Consistency Score (SCS)}

SCS measures intra-prompt stability by quantifying the average similarity of multiple responses generated from the same prompt. Given a prompt $p$ and $n$ decoded outputs $\{ y_1, y_2, \dots, y_n \}$ from the same model, we define:

\[
\mathrm{SCS}(p) = \frac{2}{n(n-1)} \sum_{1 \le i < j \le n} \mathrm{sim}(\mathbf{e}_{y_i}, \mathbf{e}_{y_j})
\]

where $\mathbf{e}_{y_i}$ denotes the Sentence-BERT embedding of response $y_i$, and $\mathrm{sim}(\cdot,\cdot)$ is cosine similarity. The final score is averaged over all prompts:

\[
\mathrm{SCS} = \frac{1}{|\mathcal{P}|} \sum_{p \in \mathcal{P}} \mathrm{SCS}(p)
\]

Higher SCS indicates more stable and deterministic generation under identical input conditions.

\subsection{Semantic Diversity Score (SDS)}

SDS quantifies inter-agent response variation when multiple agents process the same prompt. For a prompt $p$, let $\{ y_1, y_2, \dots, y_k \}$ be the responses generated by $k$ distinct agents. The diversity score is defined as:

\[
\mathrm{SDS}(p) = \frac{2}{k(k-1)} \sum_{1 \le i < j \le k} \left[ 1 - \mathrm{sim}(\mathbf{e}_{y_i}, \mathbf{e}_{y_j}) \right]
\]

where $\mathbf{e}_{y_i}$ is the Sentence-BERT embedding of response $y_i$, as before. The final SDS is computed by averaging over the full prompt set:

\[
\mathrm{SDS} = \frac{1}{|\mathcal{P}|} \sum_{p \in \mathcal{P}} \mathrm{SDS}(p)
\]

A higher SDS implies greater semantic variability across agents, which is desirable in scenarios requiring exploration and heterogeneous coordination.

\end{document}